\documentclass[letterpaper, 10 pt, conference]{ieeeconf}  

\IEEEoverridecommandlockouts                              

\usepackage[utf8]{inputenc} 
\usepackage[T1]{fontenc}    
\usepackage{hyperref}       
\usepackage{url}            
\usepackage{booktabs}       
\usepackage{amsfonts}       
\usepackage{nicefrac}       
\usepackage{microtype}      
\usepackage{graphicx}
\usepackage{amsmath} 
\usepackage{amssymb}  
\usepackage{xcolor}
\let\labelindent\relax
\usepackage{enumitem}
\usepackage{dsfont}

\title{\LARGE \bf
Tractable Joint Prediction and Planning over Discrete Behavior Modes for Urban Driving
}

\author{Adam Villaflor$^{1}$, Brian Yang$^{1}$, Huangyuan Su$^{1}$, Katerina Fragkiadaki$^{1}$, John Dolan$^{1}$, Jeff Schneider$^{1}$
\thanks{$^{1}$Carnegie Mellon University, Pittsburgh, PA, USA.
        {\tt\small \{avillafl,brianyan,huangyus,kfragki2,
        jdolan,jeff4\}@andrew.cmu.edu}}%
}

\begin{document}

\maketitle
\thispagestyle{empty}
\pagestyle{empty}

\begin{abstract}
Significant progress has been made in training multimodal trajectory forecasting models for autonomous driving.
However, effectively integrating these models with downstream planners and model-based control approaches is still an open problem.
Although these models have conventionally been evaluated for open-loop prediction, we show that they can be used to parameterize autoregressive closed-loop models without retraining.
We consider recent trajectory prediction approaches which leverage learned anchor embeddings to predict multiple trajectories, finding that these anchor embeddings can parameterize discrete and distinct modes representing high-level driving behaviors.
We propose to perform fully reactive closed-loop planning over these discrete latent modes, allowing us to tractably model the causal interactions between agents at each step.
We validate our approach on a suite of more dynamic merging scenarios, finding that our approach avoids the \textit{frozen robot problem} which is pervasive in conventional planners.
Our approach also outperforms the previous state-of-the-art in CARLA on challenging dense traffic scenarios when evaluated at realistic speeds.
\end{abstract}

\section{Introduction}
Imagine a situation where a vehicle is exiting a crowded parking lot after a football game.
Most drivers know that in bumper-to-bumper traffic, they should behave proactively by cautiously asserting themselves to make space for themselves and progress forward.
While this is intuitive for human drivers, these highly interaction-dense scenarios are a major challenge for current autonomous vehicles.

In these situations, we require a robust model of how other agents might behave, and crucially, how they might respond to our own actions.
To this end, significant progress has been made towards learning trajectory forecasting models from large datasets of urban driving logs \cite{tang2019multiple,varadarajan2022multipath++,ngiam2021scene,girgis2021autobots,casas2020implicit}.
State-of-the-art trajectory prediction models can capture the highly stochastic and multimodal distribution of outcomes in driving without needing to manually engineer complex human driving behaviors.
In this work, we aim to explore fully leveraging these learned models for model-based planning and control.

Planning and prediction are usually treated as separate modules within a conventional autonomy stack.
Usually a forecasting model will make predictions for all of the actors in the scene, and then plan open-loop against a set of possible open-loop trajectories for other agents in the scene.
While this may ensure collision avoidance, it can cause the agent to behave overly conservatively, resulting in the \textit{frozen robot problem} -- if we are highly uncertain about nearby actors, then the only perfectly safe course of action is to do nothing, which can result in deadlock.
Ideally, we should perform fully reactive closed-loop prediction and planning, such that the ego-vehicle actions directly affect the predicted behaviors of other agents, and vice versa.

Unfortunately, performing fully reactive closed-loop planning over a learned forecasting model is often computationally intractable.
The distribution over agent trajectories is highly multimodal and subject to change at every timestep, which means the space of possible outcomes grows exponentially over time.
It is also impossible to enumerate and check every possible future, since the distribution is continuous.
Existing work \cite{rhinehart2021contingencies} performs fully reactive closed-loop planning by first parameterizing agent policies as continuous latents in a normalizing flow \cite{rezende2015variational}, and then optimizing the ego-agent policy with respect to some differentiable cost function, but these approaches have previously only been shown to work at small scale with one or two other interacting agents.
We show later that our approach scales favorably on more challenging tasks.

In this work, we devise a novel approach for performing fully reactive closed-loop planning over multimodal trajectory prediction models.
We consider recent approaches to trajectory prediction that leverage learned anchor embeddings \cite{varadarajan2022multipath++,girgis2021autobots,ngiam2021scene} to predict a diverse set of trajectories.
While these approaches have previously only been evaluated for open-loop prediction, to the best of our knowledge ours is the first work to show how these models can also be used for closed-loop planning for dense urban driving.
Our contributions are as follows:
\begin{itemize} 
    \item \textbf{Closed-loop prediction with open-loop training.}
    We show how trajectory prediction models trained open-loop with learned anchor embeddings can be used to parameterize autoregressive closed-loop models without the need for retraining.
    \item \textbf{Fully reactive closed-loop planning over discrete latent modes.}
    We propose a novel planning approach which leverages discrete latent modes to do planning over a compact latent behavior space.
    Our planning approach scales linearly with the number of agents and the planning horizon, as opposed to naive search, which scales exponentially.
    This allows us to tractably model the causal interactions between agents when performing autoregressive rollouts; the predictions for other agents react to the planned ego-trajectory and vice versa.
    Similar to \cite{rhinehart2021contingencies}, this allows us to perform fully reactive closed-loop planning, but over discrete rather than continuous latent conditioned policies.
    We find that our discrete formulation ensures diverse trajectory proposals and improves downstream planning performance.
    
\end{itemize}

We validate our approach on a challenging suite of highly interactive urban driving scenarios, outperforming the demonstrator agent and several strong baselines.
Our approach also beats the previous state-of-the-art in CARLA \cite{Dosovitskiy17} on the Longest6 benchmark \cite{chitta2022transfuser} when evaluated at realistic speeds.

Our code is available at \url{https://github.com/avillaflor/P2DBM}.

\section{Related Work}
\subsection{Trajectory forecasting models for driving}

There has been a substantial amount of work towards training trajectory forecasting models for urban driving \cite{rhinehart2019precog,tang2019multiple,casas2020implicit,ngiam2021scene,girgis2021autobots,varadarajan2022multipath++}.
However, none of these papers performs evaluations on closed-loop planning tasks beyond doing log-replay on offline driving logs for short time horizons.
Most closely related to the model we use are Scene Transformer \cite{ngiam2021scene} and AutoBots \cite{girgis2021autobots}, which both use vector-based input representations and large transformer models.
Our model can support many state-of-the-art forecasting approaches, and only requires minimal modification to most in order to facilitate planning.
Note that although all of these approaches are trained to perform open-loop prediction, our approach demonstrates how they can be adapted to perform fully reactive closed-loop prediction and planning without needing to substantially modify the training procedure.

\subsection{Planning over learned forecasting models}

There is a substantial amount of literature on planning over learned forecasting models for driving.
Several prior works plan over imitation models, but do not perform closed-loop planning, instead optimizing a fixed ego-trajectory \cite{rhinehart2018deep,zeng2019end,song2020pip,liu2021deep,hu2023planning}.
Another common approach is to not model the influence of the ego-agent on other agents \cite{hardy2013contingency,zhan2016non,cui2021lookout}.
As we will demonstrate, these approaches are unable to perform proactive planning when forced into close interactions with dense traffic.
Model-based reinforcement learning approaches such as \cite{wu2021uncertainty,kamenev2022predictionnet} can theoretically leverage learned models and perform closed-loop prediction and planning for autonomous driving.
However, many of these approaches have been restricted to small state spaces and have not been shown to perform well on more difficult closed-loop driving environments.
In particular, we argue that rich vector-based state representations such as the one used in this work are key to effective prediction and planning.
\cite{huang2023gameformer} models joint interactions between agents using a game-theoretic framework and evaluates planning performance in simulation, but does not use online fully reactive closed-loop planning like our approach.

The most similar work to ours in the literature is CfO \cite{rhinehart2021contingencies}, which also does fully reactive closed-loop planning for driving in CARLA.
The tasks they consider are significantly simpler than the ones considered in this paper -- they only consider interactions with one other vehicle (we consider up to 100).
Their approach also relies on an autoregressive normalizing flow model which, to our knowledge, has not yet been scaled to larger-scale datasets, whereas our model architecture is similar to other transformer-based forecasting models in the literature which have been deployed at scale.

\subsection{Learning to drive in CARLA}
Most competing approaches in CARLA \cite{Dosovitskiy17} are based on imitation learning \cite{chen2020learning,chen2021learning,chen2022learning,chitta2022transfuser,Chitta2021ICCV}.
While most approaches focus on the visual imitation setting, our closest competitor is PlanT \cite{renz2022plant}, which also does imitation learning, but operates in a similar setup to ours, where we assume the perception problem is solved.
Imitation learning approaches rely on collecting optimal demonstrations, which can be challenging as we scale up to harder scenarios and more realistic data collection settings.
We show that unlike imitation learning, our method is able to out-perform the demonstrator on our suite of challenging urban navigation tasks.

\section{Trajectory Prediction}

\subsection{Model formulation}

In this section, we present our model architecture used for trajectory prediction.
Let $x_{t}^{a}$ denote the state of vehicle $a$ at timestep $t$ (we assume $x_{t}^{0}$ and $x_{t}^{1:A}$ are the vehicle states for the ego-vehicle and the rest of the vehicles respectively).
We also assume that $t=0$ is the current timestep and $x_{\leq T}$ refers to future vehicle states $x_{1},...x_{T}$ (ignoring the current timestep).
Our goal is to model the distribution over future vehicle states $P(x_{\leq T}| x_0, c)$ where $c$ is some arbitrary conditioning information (omitted from now on for brevity).
We can factorize this distribution autoregressively over time, as is common in conventional trajectory prediction approaches.
In this work we only model marginal (as opposed to joint) distributions over agents, so we can additionally factorize our distribution as
\begin{align}
    P(x_{\leq T}| x_0) = \prod_{t=0}^{T} \prod_{a=0}^{A} P(x_{t+1}^{a} | x_t)
\end{align}

We propose a simple model formulation which captures multimodality and is well-suited for downstream reactive planning.
We represent the high-level multimodal behavior of the agents with categorical latent variables $z_{t}^{a} \in [1,...,K]$ for each agent:
\begin{align}
\label{eq:model}
    P(x_{\leq T}| x_0) = \prod_{t=0}^{T} \prod_{a=0}^{A} P(x_{t+1}^{a} | x_t, z_{t}^{a}) P(z_{t}^{a} | x_t)
\end{align}
Following \cite{casas2020implicit}, we implicitly characterize $P(x_{t+1}^{a} | x_t, z_{t}^{a})$ using a deterministic mapping $x_{t+1}^a = f(x_t, z_{t}^a)$.
Therefore, we only require our model to learn this 1-step prediction mapping $x_{t+1}^a = f(x_t, z_{t}^a)$, and estimate $P(z_{t}^{a} | x_t)$.
In practice, we find that training our model to predict $H$ steps open-loop into the future $x_{t+1:t+H}^a = f(x_t, z_{t}^a)$ is a useful auxiliary task that improves the training and generalization of the desired 1-step prediction mapping.
Below we show how $f(x_t, z_{t}^a)$ and $P(z_{t}^a | x_t)$ can be parameterized by a transformer model.

\subsection{Network architecture}

\begin{figure}
    \vspace{10pt}
    \centering
    \includegraphics[width=.46\textwidth]{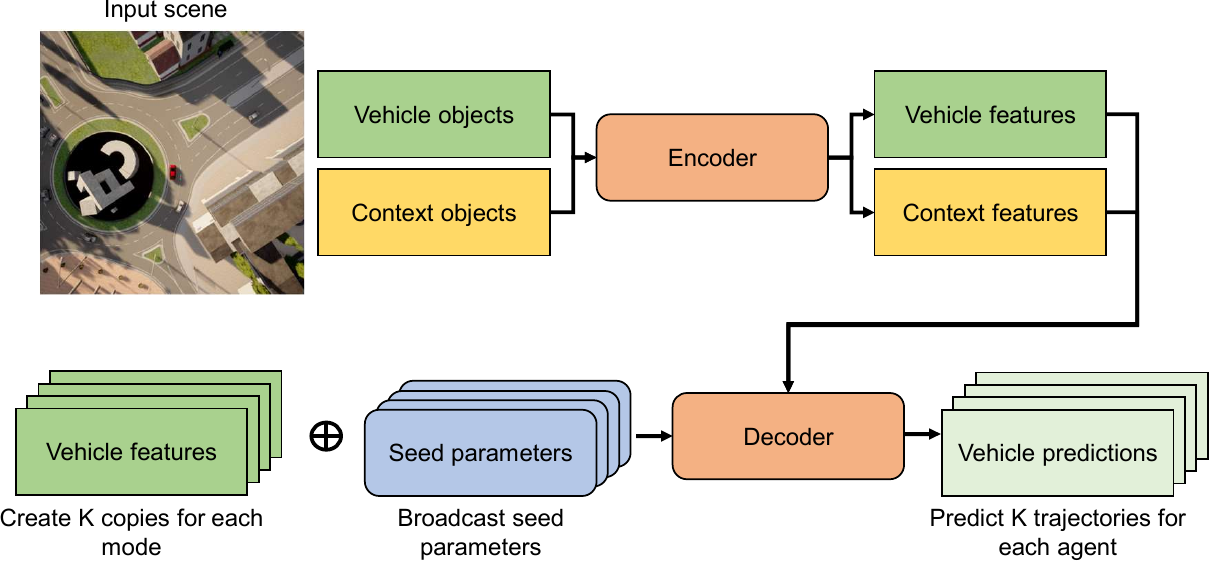}
    \caption{Forecasting model architecture.}
    \label{fig:arch}
\end{figure}

We design our architecture to predict each of the different modes for all agents independently, but all in one pass of the model. Thus, our model outputs a vector with shape $[K, A, H, 4]$ of potential future states of all agents in the scene, where $K$ is the number of modes and $A$ is the number of agents. This corresponds to $K\times A$ trajectories of length $H$, where each waypoint consists of position, heading, and speed.
The high-level structure of our model is depicted in Figure \ref{fig:arch}.

Since the number of agents and context objects in the scene is not fixed, we adopt a flexible vector-based representation of the scene.
We consider two types of entities: vehicles $\mathcal{X}$ and scene context objects $\mathcal{C}$ which consist of road points, traffic lights, pedestrians, stop signs, and goal waypoints.
Each object is represented by a feature vector that contains the relevant raw information.
For vehicles, this is relative pose, bounding box, speed, and current effective speed limit.
For road points this is relative pose, lane width, whether the point is in an intersection, and whether one can change lanes to the left and right.
For traffic lights this is relative pose, affected bounding box, and light state.
For pedestrians this is relative pose, bounding box, and speed.
For stop signs this is relative pose and affected bounding box.
For goal waypoints this is relative pose and lane width.
We assume access to all agents and context objects within 50m of the ego-vehicle up to a hard cap for each class of object.
Each object is encoded using a MLP (one for each object type) to ensure all object features are the same dimension $D$.
Then the entire set of input features is processed by an encoder to produce encoded vehicle features: $\mathcal{X}' = \textrm{Enc}(\mathcal{X}, \mathcal{C})$.

The encoder consists of stacked cross-attention layers where each vehicle cross-attends to both the vehicle features $\mathcal{X}$ and the context features $\mathcal{C}$ concatenated together.
In practice, we find that only allowing each vehicle to attend to nearby map features improves performance, so we only attend to the closest 50 map features.
Additionally, we use relative positional embeddings to maintain translational invariance.
Prior work \cite{cui2023gorela} has demonstrated that translational invariance is a useful inductive bias when performing multi-agent trajectory prediction.
When feature $u_i$ cross-attends to $u_j$, we compute query $Q_{ij}$, key $K_{ij}$, and value $V_{ij}$ as follows:
\begin{align*}
    Q_{ij} &= W^{q} u_i \\
    K_{ij} &= W^{k} (u_j + p_{ij}) \\
    V_{ij} &= W^{v} (u_j + p_{ij})
\end{align*}
where $W^{q}, W^{k}, W^{v}$ are learned projection matrices and $p_{ij}$ is a learned relative positional embedding.
All features have an associated spatial position $(x,y)$ as well as an orientation $\theta$.
To compute $p_{ij}$, we compute the pose $(x_j, y_j, \theta_j)$ of feature $u_j$ transformed to the frame of feature $u_i$ at pose $(x_i, y_i, \theta_i)$ to get a relative pose $(x_{ij}, y_{ij}, \theta_{ij})$.
We then construct a relative pose feature by concatenating $x_{ij}$ and $y_{ij}$, as well as $\sin(\theta_{ij})$ and $\cos(\theta_{ij})$.
Finally, an MLP is used to encode this relative pose feature to obtain $p_{ij}$.
Note that we do not include any absolute positional information in our features, so the network is only able to observe the relative poses of scene objects.

Following the encoder, our encoded vehicle features are of shape $[A,D]$.
We expand these features to be of shape $[K,A,D]$ so that each unique combination of mode and actor corresponds to a single $D$-dimensional feature.
Following \cite{varadarajan2022multipath++,ngiam2021scene,girgis2021autobots}, we introduce learnable anchor embeddings $M \in \mathbb{R}^{K\times D}$ for each of the $K$ modes, which are broadcasted and added to the encoded vehicle features in order to distinguish the different modes.
We learn two separate sets of anchor embeddings: one for the ego-vehicle only, and one for all other vehicles.

For each combination of agent and mode we get the query features $Q_{k}^a = \mathcal{X}'^a + M_k$ with appropriate broadcasting.
Then, we use the decoder to produce $K$ trajectory predictions for each agent: $Y_{k} = \textrm{Dec}(Q_{k}^a, \mathcal{X}')$.
The decoder consists of two alternating operations: the query features $Q_{k}^a$ cross-attend to both the encoded vehicle features $\mathcal{X}'$ and map features $\mathcal{C}$, and then the query features $Q_{k}^a$ self-attend along the $K$ modes dimension.
In other words, for each agent, the different mode features self-attend to each other.
Note that during self-attention, only features corresponding to the same agent can attend to each other.
This ensures that each agent future is decoded independently.

Finally, the processed query features are passed to a MLP.
Since each unique agent and mode has its own feature, each $D$-dimensional feature is mapped to a trajectory of length $H$, where each waypoint consists of positions, orientations, and speeds.
For each agent, we predict relative to that agent's coordinate frame, and also for each time step predict in-frame displacements rather than absolute positions, orientations, and speeds.

\subsection{Training objectives}
We train the model to predict the ground truth trajectories using a negative log-likelihood loss, where each prediction is parameterized by a Gaussian.
Note that although we train our model to parameterize a Gaussian, we always take the mean when unrolling our model.
In order to ensure our model can capture multimodal outcomes, we adopt a winner-takes-all objective \cite{narayanan2021divide} where only the closest of the $K$ predictions will backpropagate its loss.
Since we are doing marginal prediction, our winner-takes-all objective takes the best prediction for each agent separately.
Our model is also trained to output a logit for each mode to estimate the probability of that mode being used, and this is trained using a cross-entropy loss with a one-hot target for the closest of the $K$ predictions.

\section{Closed-Loop Planning}

\subsection{Planning with autoregressive rollouts}
Our objective is to find the behaviors for the ego-vehicle that will maximize the discounted sum of rewards over $T$ timesteps, where $\gamma$ is a discount factor: $\sum_{t}^{T} \gamma^{t} R(x_t, c)$.
Because the ego-vehicle's decision in the trajectory is determined by the chosen $z^0$, we can write our objective as
\begin{align}
\label{eq:opt}
    \operatorname*{argmax}_{z^{0}} \mathbb{E}[\sum_{t}^{T} \gamma^{t} R(x_t, c)]
\end{align}
As illustrated in equation \ref{eq:model}, we can sample $x_{\leq T}$ autoregressively using our transformer model.
For each timestep, we just deterministically decode $x_{t+1}^a = f(x_t, z_{t}^a)$ for our latent sample $z_t^a$. 

Given our specific transformer architecture, we can perform each autoregressive step of this procedure with just one forward pass of the model.
We generate all the multimodal predictions at once by leveraging the anchor embeddings and pick the mode prediction that corresponds to $x_{t+1}^a = f(x_t, z_{t}^a)$.
Note that even though we trained these models to make multi-step open-loop predictions, we only need to take the first step of the prediction, which is $x_{t+1}^a$.
Then, we can repeat this procedure up to the decided horizon $T$ in order to generate a sample of $x_{\leq T}$.

\subsection{Evaluating ego-modes}
Now in order to perform planning, we need to evaluate Equation \ref{eq:opt} for specific values of $z^0$.
Since the ego latents $z^0 = [z_{0}^{0}, ... z_{T-1}^{0}]$ can vary between timesteps, $z^0$ can take on $K^T$ possible values ($K$ possible ego-modes for $T$ timesteps), which is computationally expensive to fully enumerate.

Empirically, we find that keeping the latent modes consistent across time (i.e. $z_{0} = ... = z_{T-1}$) for both the ego-vehicle and surrounding vehicles leads to good performance, and thus we use this scheme in our experiments to reduce complexity.
We hypothesize that this works because the learned anchor embeddings encourage the model to learn locally consistent modes since the embeddings are shared globally, i.e., they are not state-conditioned.
Additionally, we find that conditioning on these different modes leads to meaningfully diverse behaviors over time.
Thus by keeping the latent modes consistent across time, we only need to compare $K$ ego latent modes to optimize Equation \ref{eq:opt}.

Specifically, we evaluate Equation \ref{eq:opt} for any specific $z^0$ by generating $N$ samples of $x_{\leq T}$ with the previously described procedure.
At the first time step, for each surrounding agent we sample $z_0^a \sim P(\cdot| x_0)$ and for the ego-agent we set $z_0^0$ to the ego mode we are evaluating.
Then, we continue to set $z_t = z_0$ at every subsequent timestep during autoregressive generation.
We estimate $\mathbb{E}[\sum_{t}^{T} \gamma^{t} R(x_t, c)]$ by taking the mean over these $N$ samples.
We do this for all $K$ ego modes for $z^0$, and pick the mode with the maximum expected sum of rewards. We use MPC and execute the first step by using it as a target for a PID controller, and continue to replan at every step. 

This approach enables us to perform fully reactive closed-loop planning.
Note that we are planning over latent modes rather than open-loop trajectories, which means we can effectively model the causal influence of the ego on other vehicles and vice versa.
Additionally, this allows us to be robust to different responses from other agents, since we can evaluate each candidate ego latent $z^0$ under a variable number of latent samples.

\subsection{Reward function}
Our reward function is as follows:
\begin{align*}
    R(x_t, c) &= \beta_{coll} R_{coll}(x_t, c) + \beta_{lane} R_{lane}(x_t, c) \\
    &+ \beta_{speed} R_{speed}(x_t, c) + \beta_{light} R_{light}(x_t, c) \\
    R_{coll}(x_t, c) &= -\mathds{1}_{coll} \\
    R_{lane}(x_t, c) &= 1 - \frac{\lvert \text{lateral}(x_t,c) \rvert}{0.5 \times (\text{lane width})} \\
    R_{speed}(x_t, c) &= 1 - \frac{\lvert \text{speed}(x_t) - (\text{speed limit}) \rvert}{(\text{speed limit})} \\
    R_{light}(x_t, c) &= -\mathds{1}_{red}\frac{\text{speed}(x_t)}{\text{speed limit}}
\end{align*}
where $\beta_{coll}$, $\beta_{lane}$, $\beta_{speed}$, and $\beta_{light}$ control the relative weight assigned to each reward term.
The collision term $R_{coll}$ is -1 and leads to a termination if the ego collides with another object and 0 otherwise.
The lane term $R_{lane}$ penalizes deviation from the route, where $\text{lateral}(x_t, c)$ is the lateral distance of the ego from the closest route segment.
The speed term $R_{speed}$ encourages the ego to drive at the speed limit.
The light term $R_{light}$ penalizes the ego for having non-zero speed if the light is red; the penalty is scaled by the driving speed proportional to the speed limit.

\section{Results}

\subsection{Model training and hyperparameters}
For all experiments, we train our method with 4 encoder layers consisting of a cross-attention layer followed by a feedforward layer, and 4 decoder layers consisting of a cross-attention layer, self-attention layer, and a feedforward layer.
Following \cite{xiong2020layer}, we use Pre-Layer Normalization.
For the merging scenarios, we use an embedding size of 128 for a model with approximately 1.9 million parameters.
For the CARLA Longest6 benchmark, we use an embedding size of 256 for a model with approximately 7.4 million parameters.
We train our model with AdamW \cite{AdamW} \cite{Adam} with an initial learning rate of $2e-4$ with cosine annealing to 0 as we train for 100 epochs.
We train our multimodal model with $K=8$ latent modes and use $N=8$ samples to estimate expected rewards during planning.

For both experiment settings, we use a visualization range of 50m and observe the closest vehicles by distance up to a cap of 100 vehicles.
For the merging scenarios, we set the coefficients of the cost function as $\beta_{coll} = 20$, $\beta_{lane} =0.1$, $\beta_{speed} = 1$, and $\beta_{light} =0$, as there are no traffic lights in these scenarios.
For the CARLA Longest6 benchmark, we set the coefficients of the cost function as $\beta_{coll} = 20$, $\beta_{lane} = 1$, $\beta_{speed} = 1$, and $\beta_{light} =4.$

\subsection{Merging scenarios}

\begin{figure}
    \vspace{10pt}
    \centering
    \includegraphics[width=.15\textwidth]{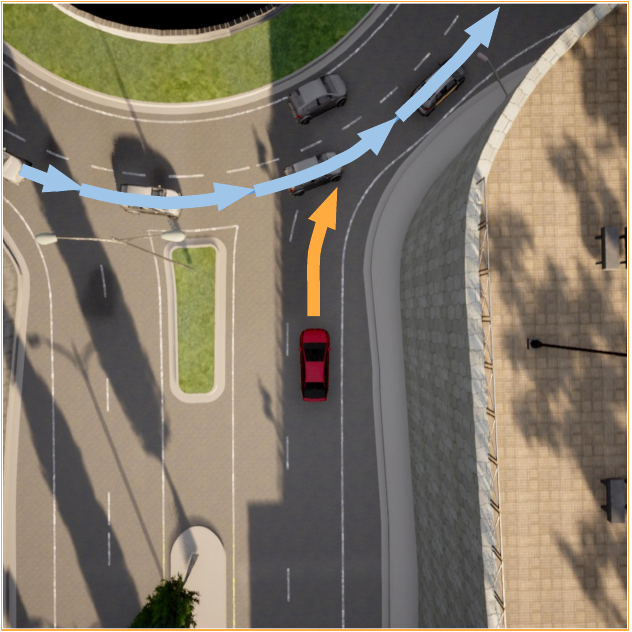}
    \includegraphics[width=.15\textwidth]{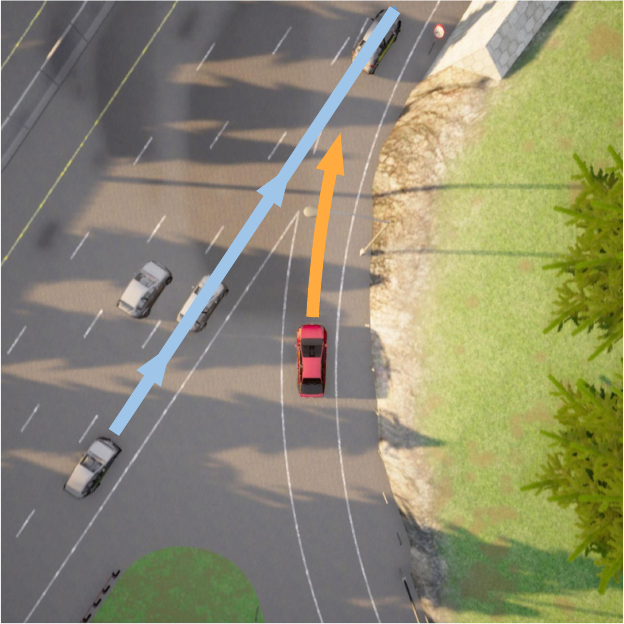}
    \includegraphics[width=.15\textwidth]{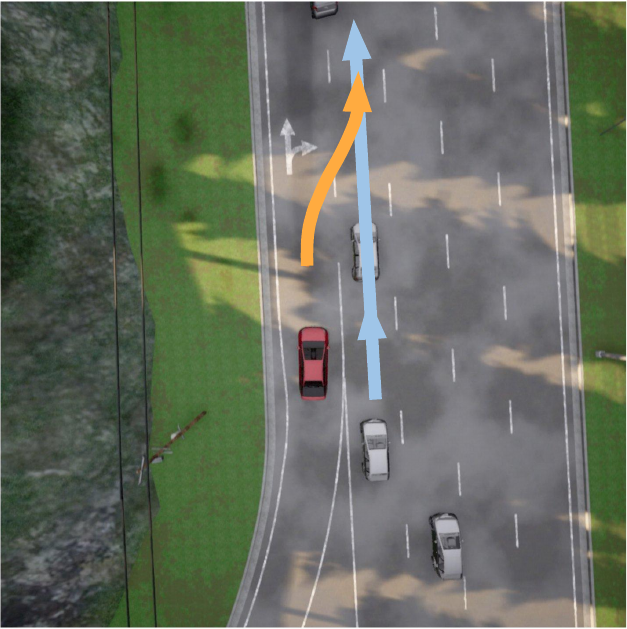}
    \caption{Merge scenario visualization. The red car is the ego-vehicle, the orange arrow indicates the desired merging behavior, and the blue arrows indicate the flow of traffic in the target lane.}
    \label{fig:carla}
\end{figure}

We design a suite of challenging urban navigation scenarios in order to evaluate whether our approach can safely perform proactive maneuvers in dense traffic.
We identify 10 distinct merging scenarios that each involve a high degree of interaction with other cars in the midst of dense urban traffic.
These other cars are controlled by CARLA's internal traffic manager, which uses an adaptive controller with randomly initialized minimum safety distances and target speeds.
In each scenario, the goal is point-to-point navigation where the goals are specified a priori.
Three of these scenarios are visualized in Figure \ref{fig:carla}.

\subsubsection{Metrics and baselines}
An episode is considered a success if the ego-vehicle reaches the goal without committing any traffic infractions; otherwise, the episode is considered a failure.
A common theme we found in prior CARLA driving benchmarks was that they rarely assessed the ego-agent's assertiveness, instead relying solely on metrics related to collisions and other traffic infractions.
As such, many of the top-performing approaches in CARLA tend to engage in overly conservative and unrealistic behaviors.
To additionally assess the ability of each approach to be proactive, we introduce an additional infraction type for static episodes, i.e., situations where the car is stopped for long periods of time.

\begin{table}
  \caption{Merge scenarios}
  \label{fig:merge}
  \centering
  \begin{tabular}{llll}
    \toprule
    \cmidrule(r){1-4}
    Approach & Success (\%) $\uparrow$ & Static (\%) $\downarrow$ & Crash (\%) $\downarrow$ \\
    \midrule
    Ours & \textbf{96.5 $\pm$ 1.8} & 2.2 $\pm$ 1.0 & \textbf{1.2 $\pm$ 0.8} \\
    Open-Loop & 78.3 $\pm$ 2.1 & 20.5 $\pm$ 2.3 & \textbf{1.2 $\pm$ 0.2} \\
    Multimodal IL & 50.2 $\pm$ 1.0 & 46.5 $\pm$ 1.5 & 3.3 $\pm$ 0.6\\
    Unimodal IL & 51.8 $\pm$ 1.6 & 46.3 $\pm$ 2. & 1.8 $\pm$ 1.0 \\
    Continuous Latents & 71.8 $\pm$ 0.8 & 24.3 $\pm$ 1.4 & 3.8 $\pm$ 0.6 \\
    \midrule
    PlanT & 67.2 $\pm$ 1.2 & 30.7 $\pm$ 0.9 & 2.2 $\pm$ 0.3 \\
    Data Policy & 68.0 & 12.5 & 19.5 \\
    Autopilot & 85.5 & \textbf{1.5} & 13.0 \\
    \bottomrule
  \end{tabular}
\end{table}

For each approach, we evaluate it on all 10 merging scenarios, each with 20 different random traffic initializations for a total of 200 episodes.
We separately report success rates, static rates, and crash rates for each approach.
On these merge scenarios, we compare to several baselines:
\begin{itemize}
    \item \textit{Multimodal imitation}.
    Same transformer model as our approach, except we execute the mode prediction with the highest predicted probability.
    \item \textit{Unimodal imitation}.
    Same transformer backbone as our approach, except we only use one mode and only make one prediction.
    \item \textit{Ours w/ open-loop planning}. 
    Same model as our proposed approach, except instead of unrolling our model autoregressively, we run our model once to generate marginal open-loop predictions for all cars. The planning procedure is the same as in the closed-loop case, except the trajectory predictions are fixed.
    \item \textit{Ours w/ continuous latent variables}.
    We train a CVAE with Gaussian latent variables with our transformer backbone and do sampling-based planning for all agents (including the ego). The number of samples is equivalent to the total number of mode combinations expanded by our approach in order to ensure fair comparison.
    \item \textit{PlanT}.
    We compare to PlanT \cite{renz2022plant}, which is an imitation learning approach that is state-of-the-art on the CARLA Longest6 benchmark.
    We use the original code implementation of the model.
    \item \textit{Data policy}. 
    The configurable heuristic policy used to collect the dataset.
    We sample random configuration parameters (e.g. planning horizon, bounding box sizes, etc.) to generate diverse demonstrations.
    \item \textit{Autopilot policy}.
    The best-performing configuration of the data collection policy.
\end{itemize}

To train all approaches, we collect a dataset of 100K samples using the data policy.
We train our models on all scenarios and share them across baselines where possible.
We report mean and standard error across 3 different model seeds.

\subsubsection{Performance analysis}

\begin{figure}
    \vspace{10pt}
    \centering
    \includegraphics[width=.15\textwidth]{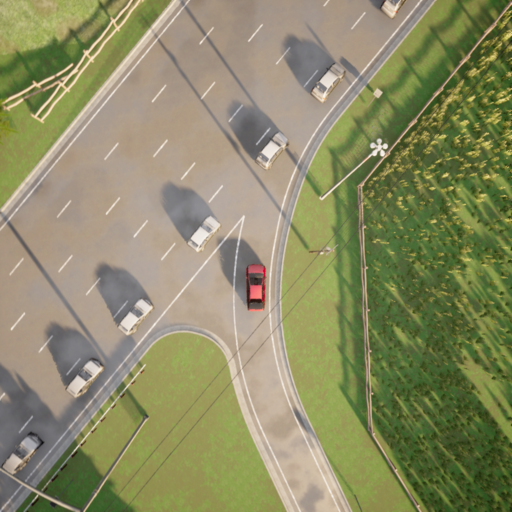}
    \includegraphics[width=.15\textwidth]{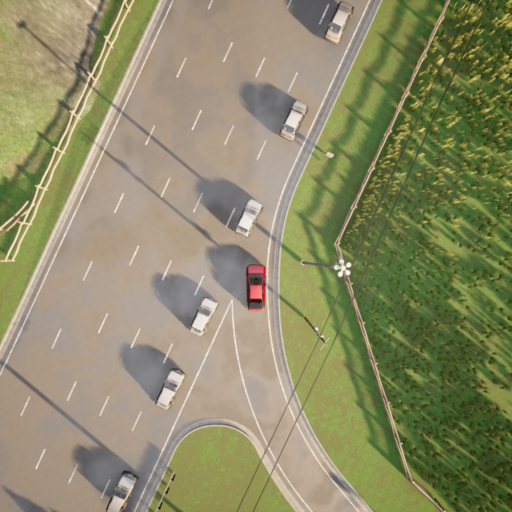}
    \includegraphics[width=.15\textwidth]{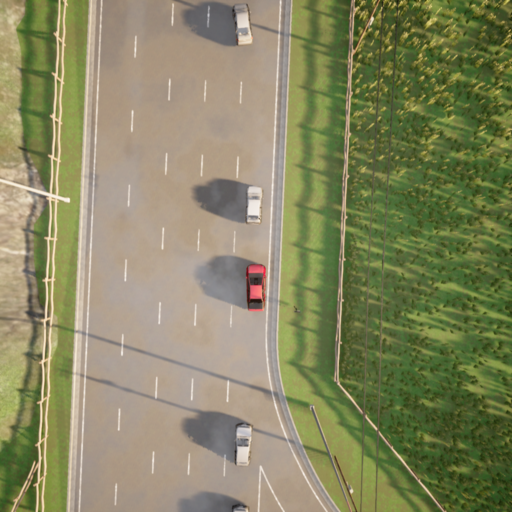} \\
    \vspace{4pt}
    \includegraphics[width=.15\textwidth]{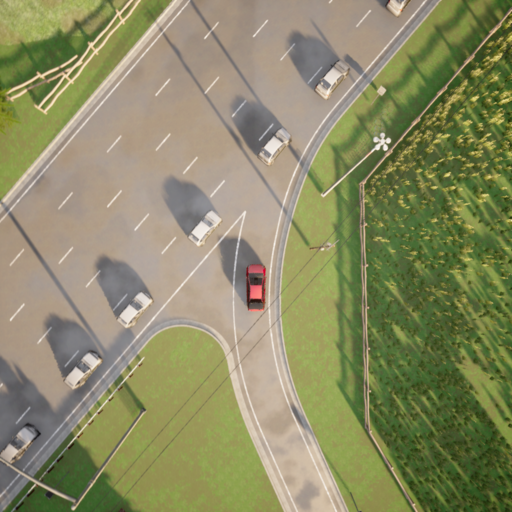}
    \includegraphics[width=.15\textwidth]{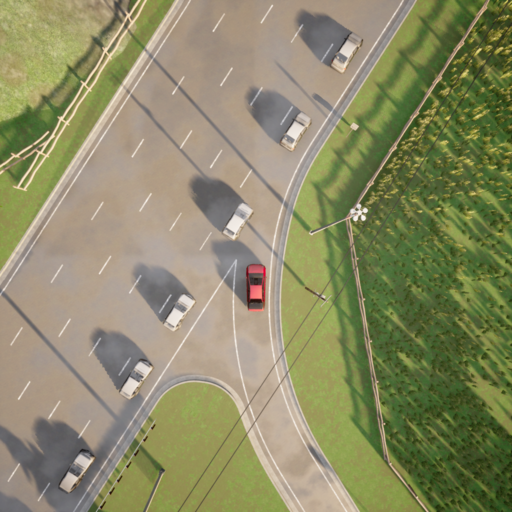}
    \includegraphics[width=.15\textwidth]{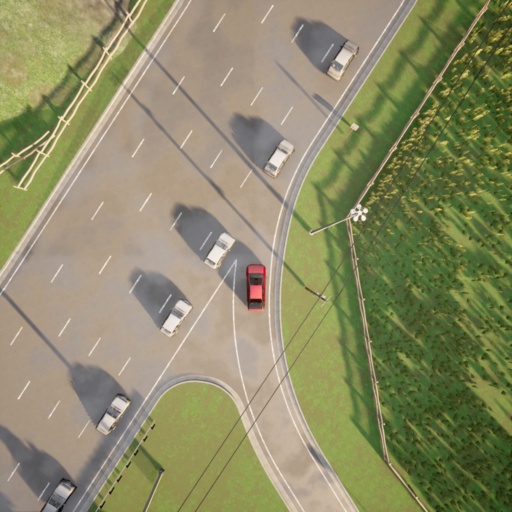}
    \caption{Qualitative example of proactive merging with closed-loop planning. The top row is from our proposed closed-loop planner and the bottom row is from the open-loop variant of our planner. Frames are in sequential order from left to right. The closed-loop planner merges proactively in front of other cars, causing the car behind to yield to the ego-agent. The open-loop planner does cannot predict that the ego will affect the behavior of other cars, so it does not attempt to merge.}
    \label{fig:openvsclosed}
\end{figure}

Table \ref{fig:merge} shows the results on the merge scenario suite.
Our approach outperforms the baselines and achieves the highest overall success rate.
Notably our method is the only learned approach which actually outperforms the best-performing heuristic policy (autopilot).
The open-loop variant of our approach performs worse than our closed-loop approach and has a higher static rate.
The open-loop planner cannot predict that other vehicles will alter their current trajectory if the ego attempts to merge and therefore behaves much more conservatively, resulting in more static episodes.
On the other hand, our closed-loop planner will predict that other vehicles will sometimes slow down to allow the ego-vehicle to merge.
Figure \ref{fig:openvsclosed} shows a qualitative comparison between the open and closed-loop planners.
The imitation policies all have low crash rates due to excessive static episodes, and generally engage in extremely conservative behavior.
This also suggests that the dominant modes in the dataset are highly cautious, even when we explicitly model multiple modes.
However, our approach is able to identify and execute the high-performance modes in the dataset.

Since our choice of discrete latents distinguishes our approach from previous work \cite{rhinehart2021contingencies}, we provide a more detailed comparison here between our approach and a CVAE baseline which uses continuous latents based on a Gaussian prior.
We suspect the main reason for the disparity in performance is that our approach is better able to predict and sample distinct trajectories.
Empirically we find that our discrete latent modes correspond to semantically different predictions that lead to a greater diversity of closed-loop predictions for both the ego-vehicle and the surrounding vehicles.
On the other hand, taking arbitrary samples from a CVAE will not necessarily cover the relevant behavior modes, especially in a limited computation setting.
Additionally, our approach explicitly enumerates distinct latent modes for the ego-vehicle independent of the predicted likelihood under the dataset, so it can more easily capture less common modes in the data.

\subsection{Longest6 benchmark scenarios}

\subsubsection{Benchmark details}
We also evaluate our approach on the CARLA Longest6 benchmark, which consists of driving on a difficult subset of the publicly available CARLA Leaderboard routes with added dense traffic.
Unlike the previous scenario suite, these routes are much longer and consist of multiple scenarios defined according to the NHTSA typology.

We found that existing state-of-the-art approaches in CARLA Leaderboard are able to avoid many of the highly interactive scenarios by simply \textit{driving significantly under the speed limit}.
For instance, pedestrians on CARLA Leaderboard are regularly spawned to cross the road in front of the ego-vehicle at a constant speed.
Many approaches avoid these scenarios by driving at a low enough speed such that the ego-vehicle will never be going fast enough to hit the spawned pedestrians.
In general, driving at low speeds substantially reduces the degree of interaction present in these scenarios.

To circumvent this loophole, we collect demonstration data with an autopilot policy driving at the speed limit.
Specifically, we use the same autopilot to collect the data that was used in \cite{renz2022plant}, but have the target speed be the current speed limit instead of a constant 4 m/s.
This significantly increases the level of interaction required to solve these routes.
In order to facilitate fair comparison with existing approaches, we retrained a state-of-the-art approach PlanT \cite{renz2022plant} using our new dataset.

\subsubsection{Metrics and baselines}
We adopt the main metric used by the CARLA Leaderboard.
Driving score (DS) is the primary metric used to compare and rank approaches in CARLA Leaderboard, and is the product of a route completion score (RC) and an infraction penalty (IP) score.
Route completion is the percentage of the route that is completed.
Infraction penalty is a multiplier which penalizes various infractions such as collisions, running red lights, etc.

For baselines, we compare to the autopilot policy, a multimodal imitation baseline trained using our transformer backbone, and PlanT \cite{renz2022plant}.
We collect 440K samples using the modified autopilot policy and use this dataset to train all approaches.
For each approach, we evaluate on all 36 routes each with 5 different random seeds for a total of 180 episodes and we report the mean over all episodes and standard error over the 5 seeds.

\subsubsection{Performance analysis}

\begin{table}
  \vspace{10pt}
  \caption{CARLA Longest6 scenarios}
  \label{fig:longest}
  \centering
  \begin{tabular}{lllll}
    \toprule
    \cmidrule(r){1-4}
    Approach & DS $\uparrow$ & RC $\uparrow$ & IP $\uparrow$ \\
    \midrule
    Ours  & \textbf{51.0 $\pm$ 1.2} & \textbf{89.2 $\pm$ 1.7} & 0.582 $\pm$ 0.010 \\ 
    Autopilot & 44.6 $\pm$ 1.6 & 84.0 $\pm$ 1.1 & 0.546 $\pm$ 0.040\\
    PlanT & 37.1 $\pm$ 1.4 & 86.1 $\pm$ 1.5 & 0.457 $\pm$ 0.017\\ 
    Multimodal IL &  48.4 $\pm$ 2.6 & 83.3 $\pm$ 2.1 & \textbf{0.603 $\pm$ 0.021} \\
    \bottomrule
  \end{tabular}
\end{table}

Table \ref{fig:longest} shows the results on the CARLA Longest6 scenarios evaluated at reasonable driving speeds.
Our closed-loop planning approach outperforms the autopilot used to collect the dataset as well as PlanT, which is the current state-of-the-art on the original Longest6 benchmark.
While the imitation learning baseline has a better infraction penalty score, it ultimately has a lower overall driving score because it often deadlocks earlier in the denser routes and avoids many of the challenging interactions.
In contrast, our approach is able to act proactively and make much more progress in these dense scenarios and achieve a superior driving score.

\section{Conclusion}
In conclusion, we propose a novel approach for performing closed-loop planning over multimodal trajectory prediction models.
Our approach is able to excel on a challenging suite of dynamic merging scenarios that require proactive planning behaviors.
Additionally, we show that our approach outperforms state-of-the-art approaches on CARLA Longest6 scenarios when evaluated at reasonable driving speeds.

\clearpage

\bibliographystyle{IEEEtran}
\bibliography{root}

\end{document}